\pdfoutput=1
\documentclass[11pt]{article}

\usepackage{acl}

\usepackage{times}
\usepackage{latexsym}

\usepackage[T1]{fontenc}
\usepackage[utf8]{inputenc}
\usepackage{multirow}

\usepackage{microtype}
\usepackage{booktabs}
\usepackage{graphicx}
\usepackage{subfigure}
\usepackage{xspace}
\usepackage{mathtools}
\usepackage{enumitem}
\usepackage{amsmath}
\usepackage[ruled,vlined,noend]{algorithm2e}
\usepackage[bottom]{footmisc}
\newcommand{\our}{\mbox{\textsc{ZeroTOP}}\xspace}

\newcommand{\Ch}{\mathrm{I2S}}
\newcommand{\Pa}{\mathrm{S2I}}
\newcommand{\NPa}{\mathrm{S2NI}}

\title{$\our$: Zero-Shot Task-Oriented Semantic Parsing using \\ Large Language Models}

\author{Dheeraj Mekala$^{1}$\thanks{\quad Work done during an internship at Microsoft Semantic Machines} \quad Jason Wolfe$^2$ \quad Subhro Roy$^2$ \\
        $^1$UC San Diego \quad  $^2$Microsoft Semantic Machines\\
        \texttt{dmekala@ucsd.edu} \quad \texttt{sminfo@microsoft.com}}
\begin{document}
\maketitle

\begin{abstract}
% Task-Oriented Semantic Parsing converts natural language utterances pertaining to a task to their respective meaning representations (programs).
We explore the use of large language models (LLMs) for zero-shot semantic parsing. Semantic parsing involves
mapping natural language utterances to task-specific meaning representations. Language models are
generally trained on the publicly available text and code and cannot be expected to directly generalize to domain-specific parsing tasks in a zero-shot setting.  
% Training task-oriented semantic parsers is often bottlenecked by expensive manual annotation that involves identifying intent of the utterance, possible slots, and filling the slots with their respective values.
%In this paper, we explore zero-shot semantic parsing i.e. parsing utterances into meaning representations using user-provided schema without any utterance-meaning representation annotations.
In this work, we propose $\our$, a zero-shot task-oriented parsing method that decomposes a semantic parsing problem into a set of abstractive and extractive question-answering (QA) problems, 
% We, therefore, decompose a semantic parsing problem into a set of abstractive and extractive question-answering (QA) problems, 
enabling us to leverage the ability of LLMs to zero-shot answer reading comprehension questions.
% We, therefore, decompose a semantic parsing problem into a set of abstractive and extractive question-answering (QA) problems, enabling us to leverage the ability of large pre-trained language models (LLMs) to zero-shot reading comprehension questions.
% Specifically, each utterance is decomposed into multiple questions pertaining to its top level intent and its respective slots.
% In this work, we propose $\our$, a zero-shot 
For each utterance, we prompt the LLM with questions corresponding to its top-level intent and a set of slots and use the LLM generations to construct
the target meaning representation.
%A significant proportion of slots are not mentioned in any given utterance and it is necessary for an LLM to abstain from prediction for these questions.
We observe that current LLMs fail to detect unanswerable questions; and as a result cannot handle questions corresponding to missing slots. 
%They often hallucinate and fail to abstain from answering questions even under constrained settings.
To address this problem, we fine-tune a language model on public QA datasets using synthetic negative samples.
% We call our method $\our$, which stands for zero-shot task-oriented parsing.
Experimental results show that our QA-based decomposition paired with the fine-tuned LLM can 
correctly parse $\approx 16$\% of utterances in the MTOP dataset without requiring any annotated data.
\end{abstract}
\section{Introduction}
% - Explain semantic parsing and zero-shot scenario.
% - LLMs perform well on datasets as shown in several NLP benchmarks.
% - However, it is not the case in semantic parsing as it requires domain knowledge.
% - In this paper, we explore zero-shot setting.
Large pre-trained language models (LLMs) ~\cite{sanh2021multitask, gpt3, chen2021evaluating, he2021debertav3} have been shown to attain reasonable zero-shot generalization on a diverse set of NLP tasks including text classification, question answering and text summarization~\cite{wang2019superglue}.
% However, zero-shot semantic parsing is generally not considered a testbed for evaluation even though annotating training data for semantic parsing task is much more expensive than other NLP tasks. 
% It requires each utterance to be annotated with a top-level intent, and its corresponding slots to be identified, and finally identified slots are filled with their corresponding values.
% \dheeraj{Explain why annotation in semantic parsing is expensive}
% In this paper, we address such massive annotation bottleneck using LLMs.
These models when prompted with natural language description or 
a common template for the task along with the input, can directly 
generate the correct output with reasonable accuracy for many natural language
processing problems.

\begin{figure}[t]
    \center
    \includegraphics[width=0.9\linewidth]{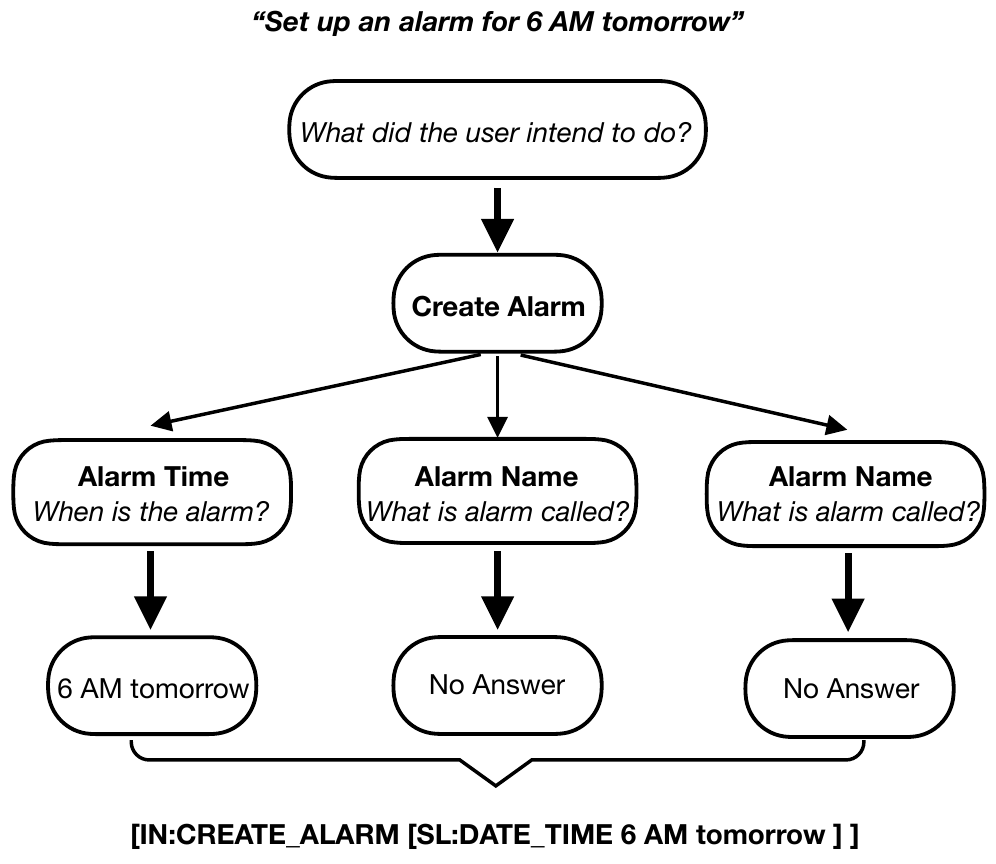}
    % \vspace{-25mm}
    \caption{
    Example of decomposing semantic parsing task into QA task.
    % \subhro{Show the meaning representation.}
    }
    \label{fig:example_qa_zsp}
\end{figure}

LLMs are generally trained on publicly available text~\cite{Liu2019RoBERTaAR, raffel2020exploring, gpt3} and code~\cite{chen2021evaluating}, and they are not expected to generalize to domain-specific semantic parsing tasks in a similar 
way. The output of semantic parsing comprises user-defined domain-specific 
functions, argument names, and syntax where the inductive bias from pre-training
is less helpful. 
Instead, we leverage the LLM's ability to zero-shot answer reading comprehension questions. 
In this work, we propose $\our$ that decomposes the semantic parsing task into one of answering a series of extractive and abstractive questions, corresponding to its top-level intent and a set of relevant slots.
For example, as shown in Figure~\ref{fig:example_qa_zsp}, the utterance is decomposed into multiple questions related to the intent (alarm creation) and slots (parameters of the alarm creation). 
The answers to these questions can be combined to infer the target meaning representation.

% \subhro{Can you restructure based on abstract? Also please look at the flow in CLAMP paper. That flow of argument will be better I think}
% Task-Oriented Semantic Parsing converts natural language utterances pertaining to a task to their respective meaning representations (programs)~\cite{mesnil2013investigation, Liu2016AttentionBasedRN, gupta-etal-2018-semantic-parsing}.
% Supervised approaches require expensive manual annotations where each utterance is annotated with an intent, and its corresponding slots are identified, and finally identified slots are filled with their corresponding values.
% % This annotation effort further increases exponentially for utterances with compositional semantics, where each slot value is annotated with its own meaning representation, producing nested programs. 
% \subhro{Semantic parsing always means compositional, so I would change this above paragraph. Leave it as
% annotating meaning representation is expensive and requires trained data curators.}

% To address this annotation bottleneck, in this paper, we explore zero-shot semantic parsing, where we aim to generate meaning representations for utterances without utterance-meaning representation pairs.
% We decompose semantic parsing task into question answering task with multiple extractive and abstractive QA problems.
% For example, as shown in Figure~\ref{fig:example_qa_zsp}, the utterance is decomposed into multiple questions related to intent and slots.

\begin{figure*}[t]
    \center
    \includegraphics[width=\linewidth]{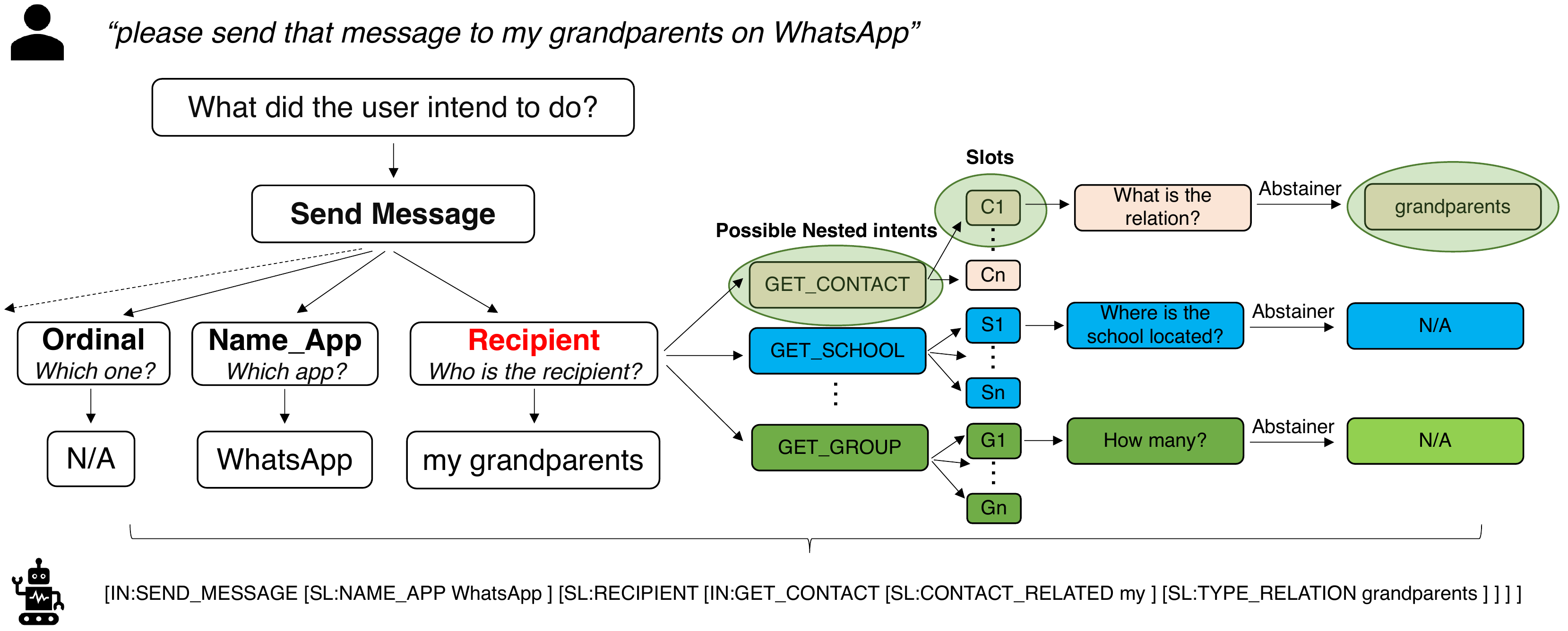}
    % \vspace{-25mm}
    \caption{
    $\our$ decomposes semantic parsing into multiple QA tasks. First, we identify top-level intent by casting it as an abstractive QA task. Next, we prompt for each slot using its corresponding question to extract their respective slot values. If a slot can accommodate nested intents (in \textcolor{red}{red}), we hierarchically prompt for nested slots. Finally, we combine all the responses to create the meaning representation.  
    }
    \label{fig:overview}
\end{figure*}

% Recently, LLMs have been shown to attain reasonable zero-shot generalization on a diverse set of tasks including QA, text summarization and this decomposition enables us to use them~\cite{sanh2021multitask, gpt3}.
As illustrated in Figure~\ref{fig:overview}, we cast top-level intent classification as an abstractive QA task. 
One intuitive way to predict intent is to constrain autoregressive LMs to search over only valid intent labels.
% prompt the constrained language model to generate the intent from possible intents.
However, LLMs are known to be biased towards predicting labels common in the pretraining data ~\cite{zhao2021calibrate}. 
Therefore, we propose to generate an intent description in an unconstrained manner from an LLM and infer the intent label most similar to the generated description.

We view slot value prediction as an extractive QA problem. Most utterances do not mention all the
slots.
% Note that, in order to predict the complete meaning representation accurately for an utterance, it is necessary for the QA model to accurately predict multiple questions pertaining to numerous slots.
% For example, in Figure~\ref{fig:example_qa_zsp}, it needs to answer 4 questions accurately.
% Moreover, there are many slots that stay unfilled for every user utterance.
For example, in the MTOP dataset, on average, only one-third of possible slots are mentioned per utterance.
It is therefore essential for the model to abstain from prediction when corresponding slots are not mentioned.
Through our analyses, we observe that most LLMs frequently hallucinate text for missing slots with high confidence, resulting in poor performance in slot value prediction.
To address this, we fine-tune a language model on a collection of public QA datasets augmented with synthetic \textit{unanswerable} samples.
We call our trained model \textit{Abstainer}, as it is capable of identifying unanswerable questions and abstaining from prediction. 
We hierarchically prompt for nested slots using the Abstainer, and infer nested intents if their corresponding slots are detected. 

% We perform experiments on the MTOP dataset.
% Through rigorous experiments and analyses, we observe that our proposed method performs better than pre-trained LLMs for zero-shot semantic parsing and 
% lays path for further exploration. 
Our contributions are summarized as follows:
\begin{itemize}[leftmargin=*,nosep]
    \item We propose $\our$ that leverages LLMs for zero-shot semantic parsing by decomposing the task into answering a series of questions.
    \item To address the LM's bias towards specific intent labels, we propose intent description generation in an unconstrained manner and then selecting most similar intent label to the description.
    \item We show that current pre-trained LLMs fail to abstain from prediction for unanswerable questions. We address this by finetuning a language model called \textit{Abstainer} on public QA datasets augmented with synthetically generated unanswerable QA pairs. Abstainer can be
    successively applied to generate nested meaning representations.
    \item Our method significantly outperforms pre-trained LLMs on the MTOP dataset.
\end{itemize}

\section{Related Work}
Large pre-trained language models are increasingly used to support semantic parsing in low-data scenarios.  
% Pre-trained and Prompt-tuned language models are utilized in various ways for semantic parsing.
~\citet{shin-etal-2021-constrained} constrain GPT-3 and BART so that the LMs generate valid canonical representations that can be mapped back to meaning representations.
~\citet{schucher-etal-2022-power, drozdov2022compositional} explore prompt tuning for semantic parsing with LLMs.
~\citet{yang-etal-2022-seqzero} decomposes canonical utterance generation into sub-clause generation and augments the generated sub-clauses into a canonical utterance.
~\citet{a-rubino-etal-2022-cross} presents a cross-schema parser for several tasks in a given vertical by augmenting schema-specific context to the input along with the utterance.
The closest work to ours is ~\citet{zhao-etal-2022-compositional} where they decompose parsing into abstractive QA tasks and aggregate answers to construct a meaning representation. These methods
assume access to some data either from the same domain, in the same format from a different domain or
synthetically generated from a synchronous grammar.
In contrast, we focus on a strict zero-shot setting where only the schema information is available along with some natural language prompts for schema entities.

%However, our method differs in terms of (1) the formulation, where they present all possible answers for a question in the input, which might not scale for large number of top-level intents and slots, and (2) evaluation, where they assume access to domain information.

Our work is also related to approaches towards zero-shot dialog state tracking using language models \cite{Gao2020, lin2021zero, lin2021leveraging}.
Specifically, ~\citet{lin2021zero} uses an Abstainer similar to ours to handle missing slots. Our method differs
in that, we focus on semantic parsing where the Abstainer needs to be applied multiple times along with 
intent detection to create nested meaning representations. Our approach does not require enumerating
choices in the prompt allowing us to handle a large number of intents and slots as found in semantic parsing 
datasets like MTOP.

\section{$\our$: Zero-Shot Task-Oriented Semantic Parsing}

%In this section, we describe the problem statement and explain our method including intent model, and slot model.

\subsection{Problem Formulation}
We focus on task-oriented parsing with hierarchical intent-slot schema.
% data which follows a hierarchical intent-slot schema.
Let $\mathcal{I} = \{\mathcal{I}_1, \mathcal{I}_2, \ldots, \mathcal{I}_n\}$ be the set of all possible top-level intents and $\mathcal{S} = \{\mathcal{S}_1, \mathcal{S}_2, \ldots, \mathcal{S}_m\}$ be the set of all possible slots. 
Each intent $\mathcal{I}_j$ has a set of slots $\mathcal{S}^j = \{\mathcal{S}^j_1, \mathcal{S}^j_2, \ldots, \mathcal{S}^j_n\}$ that can be filled.
Possible slots in an intent are represented by the intent-to-slot mapping $\Ch$: $\mathcal{I}$ $\rightarrow$ $\mathcal{P}(\mathcal{S})$, where $\mathcal{P}(\cdot)$ is the powerset operator that generates the set of all subsets. 
Similarly, the inverse slot-to-intent mapping is represented by $\Pa$: $\mathcal{S}$ $\rightarrow$ $\mathcal{I}$.
The input in a zero-shot setting contains no training data, however, we assume access to the intent-slot, slot-intent mappings $\Ch$, $\Pa$.
% and knowledge of slots that accommodate nested intents.
% To identify nested intents, we also assume knowledge of the candidate nested intents that can be accommodated by each slot, represented by the slot-to-candidate-nested-intent mapping $\NPa$: $\mathcal{S}$ $\rightarrow$ $\mathcal{P}(\mathcal{I})$.
% % To identify nested intents, we assume knowledge of the slots that accommodate nested intents and their respective possible nested intents.
% We assume that slots can nest intents to utmost one level deep i.e. nested intents cannot further have more nested intents. \subhro{I think lets not make this
% part of problem formulation; but rather a part of our solution. Also rephrase 
% ``slots can nest intents to utmost one level deep". You can directly say, ``Our
% solution assumes that the depth of output representations is at most 4". Should 
% we also show that we can try to predict deeper representations but that will hurt performance?}
Our method requires users to provide a question per slot, that is representative of their purpose.
Let $\mathcal{Q} = \{\mathcal{Q}_{\mathcal{S}_1}, \mathcal{Q}_{\mathcal{S}_2}, \ldots, \mathcal{Q}_{\mathcal{S}_k}\}$ represent respective questions for slots.
In a real-life setting, this can be easily obtained from the domain developer who designed the schema.

\subsection{Unconstrained Generation for Zero-Shot Intent Classification}
We view zero-shot intent classification as an abstractive question-answering problem and use pre-trained LLMs to answer them.
% Since multiple-choice QA requires the input to enumerate all possible answers, it is bottlenecked by the maximum length, making it infeasible for large number of intents.
% Therefore, we view intent classification as abstractive QA and use pre-trained LLMs to answer them.
One intuitive way to predict the intent is to prime the LLM with a QA prompt and then constrain the generation to search over only valid intent labels~\cite{shin-etal-2021-constrained}.
% One intuitive way to predict the intent is to \subhro{add that you have to prime LLM with a QA prompt like below and then constrain the generation} constrain LLMs to search over only valid intent labels~\cite{shin-etal-2021-constrained}.
However, LLMs are known to be biased towards certain text sequences~\cite{zhao2021calibrate} more
common in pretraining data.
For example, in the MTOP dataset, the constrained T0-3B model predicts CREATE$\_$CALL (\textit{make call}) as intent for $92\%$ of the data in the \textit{call} domain.
Therefore, we propose to first generate intent description in an unconstrained fashion and then choose the label that is most similar to the generated answer.
Specifically, we prime the LM with the below prompt and let the LM generate without any constraints.
Then, we choose the most similar label based on cosine similarity between intent label and
generated text using RoBERTa sentence similarity ~\cite{reimers-2019-sentence-bert}.
% \subhro{please check citation: this is not base roberta, but a roberta finetuned for similarity}. 
We find our proposed approach does not exhibit bias towards certain labels as noticed with constrained generation.

% RoBERTaSTS~\cite{Liu2019RoBERTaAR}(\texttt{RoBERTa-base}) to encode generated text and intent labels and choose the most similar label based on cosine similarity.
%The prompt for intent prediction is as follows:\\
\noindent\fbox{%
    % \parbox[t]{7.5cm}{
    % \parbox{\columnwidth}{
    \parbox{0.95\linewidth}{
        Answer the following question depending on the context.\\
        \textcolor{red}{\texttt{context}}: A user said, \{\textit{utterance}\}.\\
        \textcolor{red}{\texttt{question}}: What did the user intend to do?\\
        \textcolor{red}{\texttt{answer}}:
    }
}
\\

\begin{figure*}[t]
    \center
    \includegraphics[width=0.95\linewidth]{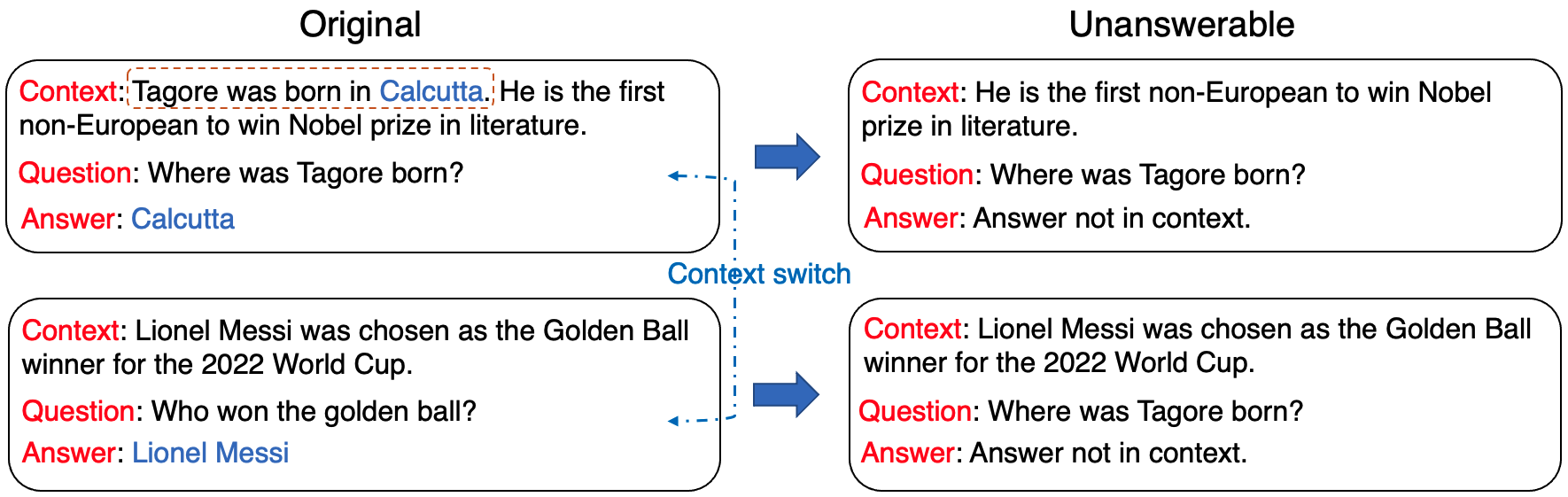}
    % \vspace{-25mm}
    \caption{
    An example demonstrating unanswerable sample generation. The first unanswerable sample is generated by removing the sentence containing the answer (highlighted with dashed lines) from the context. 
    The second unanswerable sample is generated by swapping the context of two original samples. 
    % \subhro{Make it a latex table with 3 columns. It will look cleaner.}
    % \dheeraj{I think the space won't be sufficient for a latex table.}
    % An unanswerable sample is generated by removing the sentence containing the answer (highlighted with dashed lines) from the context.\todo{Add figure for second way of unanswerable sample generation}
    }
    \label{fig:example_unans}
\end{figure*}

\begin{algorithm}[t]
\caption{$\our$: Our proposed Zero-shot semantic parsing method.}\label{algorithm:1}
\SetAlgoLined
\small
  \textbf{Input:} Set of intents $\mathcal{I}$, Set of slots $\mathcal{S}$, Slot questions $\mathcal{Q}$, intent-to-slot mapping $\Ch$, slot-to-intent mapping $\Pa$, slot-to-candidate-nested-intent mapping $\NPa$, Intent-model $\mathcal{M}_I$, Abstainer $\mathcal{M}_{abs}$, and Utterance $u$\\
  \textbf{Output:} Predicted meaning representations $\mathrm{MR}$ \\
  % $\mathrm{MR}$ = []\\
  % \subhro{remove iteration over utterances, use single utterance as input}
  % \For{{\bf utterance} $u \in \mathcal{U}$ }{
        $\mathrm{intent}$ = $\mathcal{M}_I(u)$\\
        $\mathrm{slotValues}$ = \{\}\\
        \For{{\bf slot} $\mathcal{S}_i \in \Ch(\mathrm{intent})$ }{
            $\mathrm{slotValues}[\mathcal{S}_i]$ = $\mathcal{M}_{abs}(u, \mathcal{Q}_{\mathcal{S}_i})$\\
            \For{{\bf candidate N.intent} $\mathcal{I}_j \in \NPa(\mathcal{S}_i)$ }{
                \For{{\bf slot} $\mathcal{S}_j \in \Ch(\mathrm{\mathcal{I}_j})$ }{
                    \If{$\mathcal{M}_{abs}(
                    \mathrm{slotValues}[\mathcal{S}_i], \mathcal{Q}_{\mathcal{S}_j})$ is not NONE}{
                        Update $\mathrm{slotValues}[\mathcal{S}_i]$ with nested intent $\mathcal{I}_j$, $\mathcal{M}_{abs}(
                    \mathrm{slotValues}[\mathcal{S}_i], \mathcal{Q}_{\mathcal{S}_j})$\\
                    }
                }
            }
            % \If{$\mathcal{S}_i$ accommodates nested intents}{
            %     Query nested intents\\
                
            %     Update $\mathrm{slotValues}[\mathcal{S}_i]$ with nested intents\\
            % }
        }
        $\mathrm{MR}$ = Construct representation with $\mathrm{intent}$, $\mathrm{slotValues}$\\
        % $\mathrm{MR}$ = $\mathrm{MR}$ $\cup$ $\mathrm{rep}$\\
    % }
  \textbf{Return} $\mathrm{MR}$\\
\end{algorithm}

\subsection{Leveraging QA datasets for Slot Value Prediction}
Slot value prediction involves extracting phrases for a slot from the user utterance. 
Thus, we cast this as an extractive QA problem.
All slots might not be mentioned in an input utterance. 
% Moreover, it is necessary to accurately predict all the slot questions for the resulting meaning representation to be correct, which requires the QA model 
% Moreover, for predicting the complete meaning representation accurately, it is necessary to accurately predict all the slot questions, which 
The QA model needs to abstain from prediction for the corresponding questions for the missing slots. % frequently with high precision. 
% to abstain from prediction frequently with high precision.
To analyze the abstaining capability of pre-trained QA models, we consider some of the top-performing zero-shot models T0-3B~\cite{sanh2021multitask}, GPT-3~\cite{gpt3}, and Codex~\cite{chen2021evaluating} and experiment on a 500 sample subset of unanswerable questions from the SQuAD dataset~\cite{rajpurkar-etal-2018-know}.
We constrain the models to generate either from the context or from a set of manually created phrases indicating that the question cannot be answered.
We use the prompts and phrases mentioned in ~\citet{sanh2021multitask, gpt3, chen2021evaluating} and observe the accuracy of all models to be $< 5\%$.
% To analyze the abstaining capability of pre-trained QA models, we consider state-of-the-art zero-shot model T0-3B~\cite{sanh2021multitask}, and experiment on QA datasets.
% We also experiment with constrained T0-3B, where the model is constrained to generate either from the context or from a set of manually created phrases indicating that the question cannot be answered. 
% of ``\textit{Answer not in context}''. \subhro{We have to run experiments with GPT-3 / Codex}
% The accuracy on answerable and unanswerable questions in the test sets are shown in Figure\todo{finish this}.
We notice that the LLMs frequently hallucinate and generate answers for unanswerable questions.
In our experiments section~\ref{conf}, we also consider a log-likelihood-based threshold for abstaining and we show that this threshold is difficult to tune using public QA datasets.
% \subhro{Please also add that we also consider using log likelihood of the 
% model to detect unanswerable questions. You can give a summary of your findings and point
% to experiments section}

To address this challenge, we leverage multiple publicly available extractive and abstractive QA datasets\footnote{The QA datasets details are mentioned in Appendix~\ref{app:qa_datasets}} to train \textit{Abstainer}, a QA model capable of abstaining from prediction.
Specifically, we generate synthetic unanswerable training samples by modifying
existing QA data, and train a QA model jointly on the existing datasets and the synthetic unanswerable questions.
% , augment them to the QA datasets and train a QA model on augmented data.
For every (question, answer, context) triplet, we generate synthetic unanswerable questions by either (1) removing the sentence containing the answer span from the context, or (2) randomly sampling a context that doesn't have the same question.
% For example, as shown in Figure~\ref{fig:example_unans}, we remove the sentence ``\textit{Tagore was born in Calcutta}'' containing the answer \textit{Calcutta} from the context.
For example, as shown in Figure~\ref{fig:example_unans}, we generate the first unanswerable sample by removing the sentence ``\textit{Tagore was born in Calcutta}'' containing the answer \textit{Calcutta} from the context and second unanswerable sample by switching the context of two original answerable samples.
We augment these synthetic unanswerable training samples to the existing datasets and train our Abstainer.
After training the Abstainer, we prompt it for each slot with its corresponding question for slot value prediction. The prompt has the following format:\\

\noindent\fbox{%
    % \parbox[t]{7.5cm}{
    % \parbox{\columnwidth}{
    \parbox{0.95\linewidth}{
        Answer the following question depending on the context.\\
        \textcolor{red}{\texttt{context}}: A user said, \{\textit{utterance}\}.\\
        \textcolor{red}{\texttt{question}}: \{\textit{slot question}\}\\
        \textcolor{red}{\texttt{answer}}:
    }
}
\\

\paragraph{Nested Intents} 
To identify nested intents, we assume knowledge of the candidate nested intents that can be accommodated by each slot, represented by the slot-to-candidate-nested-intent mapping $\NPa$: $\mathcal{S}$ $\rightarrow$ $\mathcal{P}(\mathcal{I})$.
% To identify nested intents, we assume knowledge of the slots that accommodate nested intents and their respective possible nested intents.
Our method assumes that the depth of output representations is at most 4 i.e. nested intents cannot further have more nested intents.
% We assume that slots can nest intents to utmost one level deep i.e. nested intents cannot further have more nested intents. 
% \subhro{Should we also show that we can try to predict deeper representations but that will hurt performance?}
% \dheeraj{This is a good point, but I guess we don't have enough time for our experiments to finish by then. I think we can keep on going deep until the Abstainer abstains for all slots. This will be a good experiment to do.}

Since we assume access to slots that accommodate nested intents, we query for possible nested intents for all such slots from their respective slot values.
One intuitive way is to prompt the LLM for nested intent with the intent prediction prompt.
However, our unconstrained generation-based intent model would predict many false positive nested intents.
We instead use Abstainer to prompt for their respective slots.
If any slot value is identified, we consider its corresponding intent via slot-to-intent mapping $\Pa$ to be present as well. 

\subsection{$\our$: Putting it all together}
The pseudo-code of the proposed pipeline is mentioned in Algorithm-\ref{algorithm:1}.
$\our$ employs a top-down, greedy prompting strategy, where we first prompt for intent and then, their respective slots.
First, we obtain the top-level intent using the intent model.
Based on the predicted intent, we prime the Abstainer for corresponding slots using their respective questions as prompts, to identify their slot values.
For each identified slot value, we prompt the Abstainer for slots
of candidate nested intents. We use the same prompt format for this
step with the identified slot value now considered as the input utterance. 
% Moreover, for each slot, we check for nested intents provided by the slot-to-candidate-nested-intent mapping $\NPa$. 
% We next treat each slot value to be a new utterance, and prompt the Abstainer for the slots of candidate nested intents in this utterance.  
% by employing Abstainer to prompt for their respective slots in the identified slot values.
Finally, we combine predicted intent, identified slot values, and nested intents to create the meaning representation.
We illustrate the pipeline with an example in Figure~\ref{fig:overview}.

% \subhro{
% % There is no text which explain the algorithm. So to understand the full pipeline, readers will have to go through the algorithm. I think we need a subsection
% % where we talk about how you put together 3.2 and 3.3. You can describe the greedy
% % approach. 
% Say what are the alternatives and that you empirically show they are 
% worse. Refer to the experiments section. You can also have this section right after
% 3.1.}
% \dheeraj{Do you think this is a better place to present alternatives like beam-search? I think it's better not to because we are describing our method here. We can discuss alternatives later in the analysis section in exps? Please let me know your thoughts.}
% \paragraph{Post-processing} We post-process blah blah blah \todo{finish this}
\section{Experiments}
In this section, we evaluate $\our$ against multiple pre-trained LLMs under zero-shot setting.
% on semantic parsing datasets.

\begin{table}[t]
    \center
    % \scalebox{1}{
    \small
    \begin{tabular}{c c c c}
        \toprule
        {\textbf{Dataset}} & {\textbf{\# Samples}} & {\textbf{\# Intents}} & {\textbf{\# Slots}} \\
        \midrule
        MTOP-En & 4386 & 113 & 74 \\
        \bottomrule
    \end{tabular}
    % }
    \caption{Relevant statistics of the dataset used in our experiments.}
    \label{tbl:datastats}
\end{table}

\subsection{Datasets}
We experiment on the English language subset of MTOP~\cite{li-etal-2021-mtop} dataset.
MTOP is a multilingual task-oriented semantic parsing dataset comprising data from 6 languages and 11 domains.
% We assume no access to domain information \subhro{you have some domain information: I2S, and slot questions} and evaluate on samples belonging to all top-level intents together.
The dataset details are mentioned in Table~\ref{tbl:datastats}.
On average, each intent has 3.6 slots and only one-third of possible slots are filled for an utterance.

\subsection{Baselines}
% We compare with state-of-the-art zero-shot LLM T0-3B as both intent and slot models.
% As an intent model, T0-3B is constrained to search over only valid intent labels. 
% As a slot model, it is constrained to generate either from the user utterance or from 
% a set of phrases indicating that the question cannot be answered. We use the same set of phrases 
% as earlier work~\citet{sanh2021multitask}. % specifying that the question is unanswerable.
% We also consider an ablated version where RoBERTa-base similarity model alone is used to predict intent.
We compare $\our$ with high-performing zero-shot LLMs as both intent and slot models. 
We consider the following models for zero-shot intent prediction:
% The compared intent and slot models are described below.
% The compared intent models are \subhro{The phrase "compared NP" does not sound right. Say "We consider the following models for zero shot intent prediction."}
\begin{enumerate}[leftmargin=*,nosep]
    \item \textbf{T0-3B, GPT-3, Codex constrained} are pre-trained T0-3B, GPT-3, Codex models that are primed with intent generation prompt and the output is constrained to search over valid intent labels.
    \item \textbf{T0-3B constrained calibrated} is same as above T0-3B constrained that is calibrated for intent labels following ~\citet{zhao2021calibrate}. Specifically, we estimate the model's bias towards each intent label by prompting with a content-free test input such as "N/A" and fit calibration parameters accordingly to cause the prediction for this input to be uniform across answers.
    \item \textbf{RoBERTa-base similarity} is an ablated version of $\our$ where we assign intent labels based only on their similarity with user utterance using the RoBERTa sentence transformer (\verb|stsb-roberta-base|).
    \item \textbf{$\our$-Intent} is our proposed intent prediction method that generates intents in an unconstrained fashion by priming T0-3B with intent generation prompt and assigns labels based on their similarity with intent labels using the RoBERTa sentence transformer (\verb|stsb-roberta-base|).
\end{enumerate}

We consider the following models for slot prediction:
\begin{enumerate}[leftmargin=*,nosep]
    \item \textbf{GPT-3, Codex, T0-3B constrained} are pre-trained GPT-3, Codex, T0-3B models that are primed with the question corresponding to slot and the output is constrained to be either from the user utterance or from a set of phrases indicating that the question cannot be answered. 
    We use their corresponding phrases mentioned in ~\citet{gpt3, chen2021evaluating, sanh2021multitask}. We use the OpenAI API \texttt{text-davinci-001} for GPT-3
    and \texttt{code-davinci-002} for Codex.
    % We use the same set of phrases as in~\citet{sanh2021multitask}.
    % \item \textbf{Codex constrained}
    % \item \textbf{T0-3B constrained} is a pre-trained T0-3B model that is constrained to generate either from the user utterance or from a set of phrases indicating that the question cannot be answered. We use the same set of phrases as in~\citet{sanh2021multitask}.
    \item \textbf{Abstainer} is our finetuned T0-3B model that abstains from prediction.
    It is primed with the question corresponding to the slot and the output is constrained to be either from the user utterance or from the set of phrases from \citet{sanh2021multitask} indicating that the question is unanswerable. 
\end{enumerate}
% \subhro{Didn't you also try GPT-3 / Codex for this? Basically using those models
% with the same prompt?}
% \subhro{Make some bullet points where you describe in more detail each baseline}

\begin{table}[t]
    \center
    % \scalebox{0.99}{
    \small
    \begin{tabular}{l c}
        \toprule
        {\textbf{Intent Model}} & {\textbf{Accuracy(\%)}} \\
        \midrule
        T0-3B constrained & $34.02$ \\
        T0-3B constrained calibrated & $36.64$ \\
        GPT-3 constrained & $40.44$ \\
        Codex constrained & $48.02$ \\
        RoBERTa-base similarity & $47.14$ \\
        $\our$-Intent & $\mathbf{49.58}$ \\
        \bottomrule
    \end{tabular}
    % }
    \caption{Top-level intent classification results. Accuracy is used as the evaluation metric.}
    \label{tbl:intent}
\end{table}

% \begin{table}[t]
%     \center
%     \scalebox{0.81}{
%     \begin{tabular}{l c c}
%         \toprule
%         {\textbf{Intent Model}} & {\textbf{Slot Model}} & {\textbf{Acc(\%)}} \\
%         \midrule
%         % GPT-3 constrained & GPT-3 constrained & \\
%         % Codex constrained & Codex constrained & \\
%         T0-3B constrained & T0-3B constrained & $2.42$ \\
%         RoBERTa-base similarity & T0-3B constrained & $3.88$ \\
%         $\our$-Intent & Abstainer & $\mathbf{15.89}$\\
%         \bottomrule
%     \end{tabular}
%     }
%     \caption{Complete meaning representation match evaluation. Accuracy is used as the evaluation metric. \subhro{Whats the difference between this and Figure 4? Note: The above combinations don't prove Abstainer helps.}}
%     \label{tbl:full_results}
% \end{table}

\begin{table}[t]
    \center
    \scalebox{0.81}{
    \begin{tabular}{l l r}
        \toprule
        {\textbf{Intent Model}} & {\textbf{Slot Model}} & {\textbf{Acc(\%)}} \\
        \midrule
        GPT-3 constrained & GPT-3 constrained & $3.00^\textbf{*}$ \\
        \midrule
        Codex constrained & Codex constrained & $5.40^\textbf{*}$ \\
        \midrule
        \multirow{2}{*}{T0-3B constrained} & T0-3B constrained & $2.42$\\
        & Abstainer & $11.81$\\
        \midrule
        \multirow{2}{*}{RoBERTa-base similarity} & T0-3B constrained & $3.88$\\
        & Abstainer & $12.90$\\
        \midrule
        \multirow{2}{*}{$\our$-Intent} & T0-3B constrained & $4.10$\\
        & Abstainer & $\mathbf{15.89}$\\
        \bottomrule
    \end{tabular}
    }
    \caption{Complete meaning representation match evaluation. Accuracy is used as the evaluation metric. To limit API cost, we limit GPT-3 and Codex evaluation on 
    a 500-example subset, and hence their results are not directly comparable.}
    \label{tbl:full_results}
\end{table}

\subsection{Experiment Settings}
We evaluate on the zero-shot setting, therefore we have no training data.
We manually create questions for slots $\mathcal{Q}$ by looking at one example per slot. 
% to design a question that is representative of the slot.
For training Abstainer, we fine-tune T0-3B on the extractive and abstractive QA datasets for $1$ epoch with a constant learning rate of $10^{-4}$.
The Abstainer is fine-tuned on $411732$ answerable and $435898$ unanswerable samples.
The batch size is $32$ and each batch contains an equal number of answerable and unanswerable samples.
We used 8 $\times$ NVIDIA Tesla V100 for our experiments.
We use complete meaning representation match accuracy as the performance metric.

\subsection{Results and Discussion}
We present intent classification results in Table~\ref{tbl:intent} and complete meaning representation evaluation results in Table~\ref{tbl:full_results}.

From Table~\ref{tbl:intent}, we observe that \our-Intent performs significantly better than constrained T0-3B, GPT-3, and Codex for zero-shot intent classification.
We found that constrained T0-3B is biased towards certain labels.
For example, it predicts CREATE$\_$CALL (\textit{make call}), SEND$\_$Message (\textit{send message}), and CREATE$\_$REMINDER (\textit{create reminder}), as intent for more than $90\%$ of the data in \textit{call}, \textit{message}, \textit{reminder} domains.
Our proposed unconstrained formulation lets the model freely express the intent and, computing similarity later with the intent labels addresses this bias. 

As shown in Table~\ref{tbl:full_results}, the combination of $\our$-Intent model and Abstainer demonstrates superior performance than % the constrained T0-3B.
alternative combinations. 
% Although $\our$ performs better than constrained GPT-3 and Codex, they are not directly comparable as they are obtained on a 500-sample subset of the test set due to its cost.
We observe that T0-3B, GPT-3 and Codex all fail to abstain frequently. 
% As a case study, we analyze the outputs of constrained T0-3B. 
The T0-3B model abstains only for $38\%$ of unanswerable slot questions whereas our Abstainer is able to abstain for $89\%$ of the unanswerable questions.
As a result, we observe a notable performance gain by plugging in Abstainer as the slot model for each intent model baseline.
This abstaining capability compounded with unconstrained generation-based intent prediction results in a significant performance gain.
% \begin{figure}[t]
%     \center
%     \includegraphics[width=0.9\linewidth]{figures/abstainer_significance.pdf}
%     % \vspace{-25mm}
%     \caption{
%     We plug in Abstainer as the slot model for each intent baseline and it results in significant improvement in performance with respect to constrained T0-3B as the slot model. 
%     % \subhro{Show the meaning representation.}
%     % \subhro{Can't read the labels}
%     \subhro{if you reduce the space between x, y axis label ("Accuracy", "Intent label") and graph, it will look
%     bigger while capturing the same space. Better still: make it a latex table.
%     Its unclear what is an intent baseline.}
%     }
%     \dheeraj{I agree with your space point. Regarding your plot vs table point, it feels like the plot makes the results look more impressive. What do you think?}
%     \label{fig:sig_abs}
% \end{figure}

% \subsection{Significance of Abstainer}
% We consider Abstainer as the slot model for each intent model baseline and present complete meaning representation match accuracy on the MTOP dataset in Figure~\ref{fig:sig_abs}.
% From the figure, we can observe a notable performance gain by plugging in Abstainer as the slot model, demonstrating its significance.

\subsection{Confidence score-based Abstainer study}
\label{conf}
\begin{figure}[t]
    \centering
    \subfigure[F1 score vs NLL threshold on MTOP dataset]{
        \includegraphics[width=0.8\linewidth]{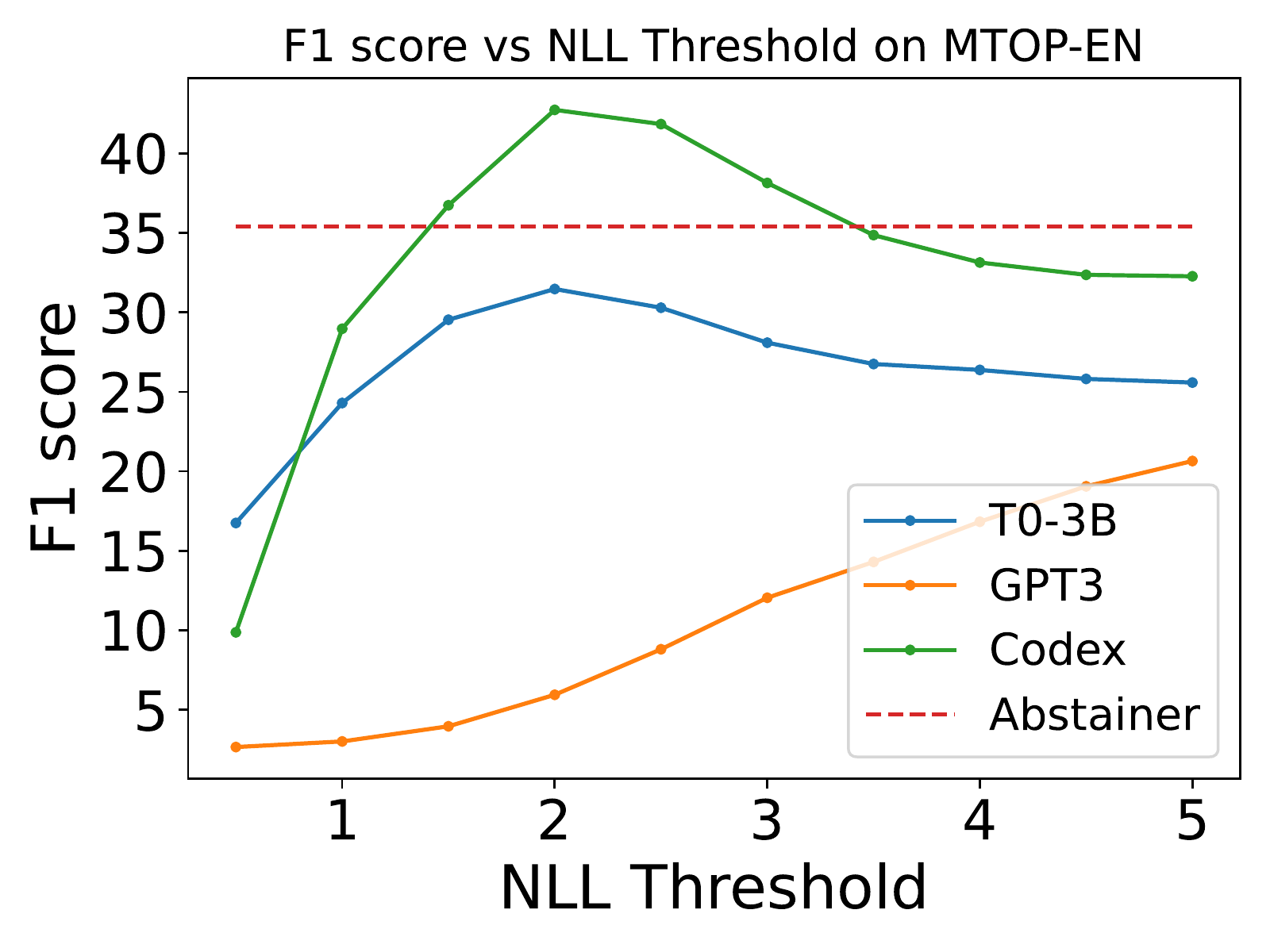}
        \label{fig:f1_mtop}
    }
    \subfigure[F1 score vs NLL threshold on SQuAD dataset]{
        \includegraphics[width=0.8\linewidth]{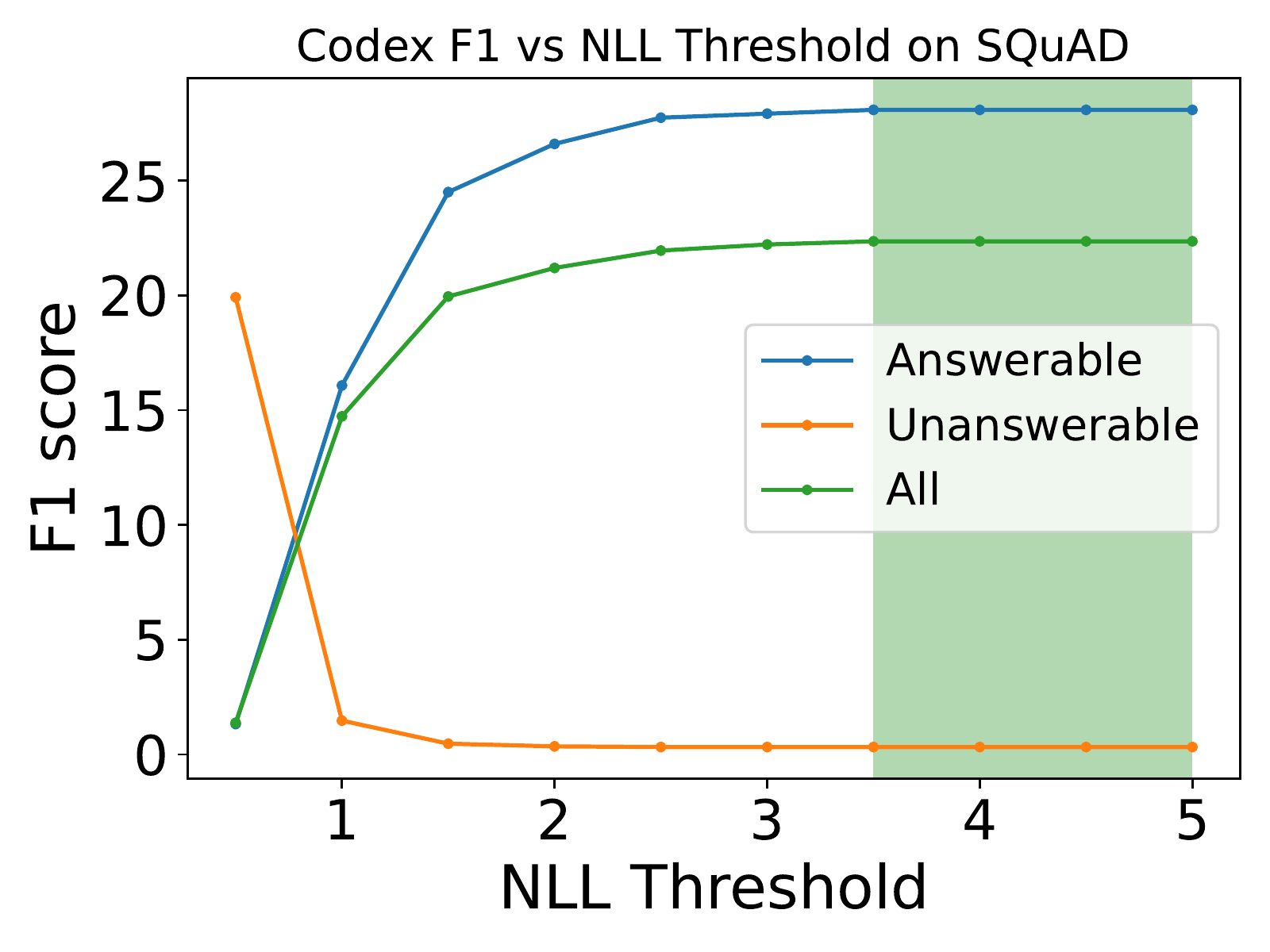}
        \label{fig:f1_squad_codex}
    }
    \vspace{-3mm}
    \caption{We consider negative log-likelihood (NLL) as a confidence score and vary the threshold to abstain from prediction and plot F1 scores on the MTOP dataset. We show that this NLL threshold is difficult to tune using public QA datasets such as SQuAD as performance on answerable and unanswerable subsets is mutually exclusive.
    % \subhro{Still not readable. You can make it one after the other, or make it figure*}
    % \subhro{Please try to make the captions here and elsewhere self explanatory. Ideally readers won't have to look anywhere else to understand the takeaways.}
    }
    \vspace{-5mm}
\end{figure}

% \begin{figure}[t]
%     \begin{subfigure}[b]
%         \includegraphics[width=1\linewidth]{figures/f1.pdf}
%         \label{fig:f1_mtop}
%         \caption{F1 score vs NLL threshold on MTOP dataset}
%     \end{subfigure}

%     \begin{subfigure}[b]
%         \includegraphics[width=1\linewidth]{figures/codex_squadf1.pdf}
%         \label{fig:f1_squad_codex}
%         \caption{F1 score vs NLL threshold on SQuAD dataset}
%     \end{subfigure}
%     % \subfigure[F1 score vs NLL threshold on MTOP dataset]{
%     %     \includegraphics[width=0.47\linewidth]{figures/f1.pdf}
%     %     \label{fig:f1_mtop}
%     % }
%     % \subfigure[F1 score vs NLL threshold on SQuAD dataset]{
%     %     \includegraphics[width=0.47\linewidth]{figures/codex_squadf1.pdf}
%     %     \label{fig:f1_squad_codex}
%     % }
%     \vspace{-3mm}
%     \caption{We consider negative log likelihood as confidence score and vary the threshold to abstain from prediction and plot F1 scores on MTOP dataset. We show that this threshold is difficult to tune using public QA datasets such as SQuAD.
%     \subhro{Still not readable. You can make it one after the other, or make it figure*}
%     \subhro{Please try to make the captions here and elsewhere self explanatory. Ideally readers won't have to look anywhere else to understand the takeaways.}
%     }
%     \vspace{-5mm}
% \end{figure}

We can alternatively have LLMs abstain from prediction based on a confidence score based threshold.
% Pre-trained LLMs can also abstain from prediction using a confidence score based threshold. 
We consider negative log likelihood (NLL) of the predicted slot value as the confidence score and abstain from prediction if it is greater than the threshold.
We experiment on slot value prediction task with T0-3B, Codex, and GPT3 as LLMs and plot macro F1 scores for multiple NLL thresholds on a randomly sampled subset of 500 samples from MTOP dataset in Figure~\ref{fig:f1_mtop}.
Specifically, we consider the gold intent of each sample and prime LLM for extracting slot values for each slot of the gold intent.
We consider F1 score as the metric due to the label imbalance across possible slot values.
We present the F1-score of the Abstainer for reference.
First, we observe that Abstainer is significantly better than T0-3B and GPT3 for all confidence thresholds.
Second, we notice that there is no threshold that consistently results in good performance for all LLMs, which implies that this has to be individually tuned for each LLM.
Finally, we observe Codex performs better than Abstainer for some thresholds.
% As our problem setting includes no annotated data, we aim to tune the threshold for Codex using public QA datasets.
As our problem setting includes no annotated data, we investigate whether we can infer
the optimal threshold for Codex using public QA datasets.
Specifically, we consider 500 answerable and 500 unanswerable QA pairs from SQuAD dataset and plot F1 scores with a range of confidence thresholds in Figure~\ref{fig:f1_squad_codex}.
We can observe that the performance on answerable and unanswerable subsets is mutually exclusive i.e. there is no threshold where the performance on both answerable and unanswerable subsets is high.
The range of thresholds that result in the best performance on the whole set (highlighted in green) does not transfer to MTOP and is achieved at the cost of unanswerable set where the F1 score is less than 5\%.
% It is possible that we can infer the optimal threshold for Codex using a larger 
% dataset. We did not explore this due to API cost limitations.
Given the difficulty in tuning threshold and the API costs of Codex, we believe using Abstainer as the slot model to be a better choice.
% \subhro{Confused by this section. 4(b) actually gets some part of the threshold
% which will give good performance in MTOP. Why can't we then use it?}
% Therefore, it is difficult to tune this threshold using 1000 samples. 
% Given the pricing of Codex and GPT3, we believe Abstainer is a better choice.

\subsection{Annotation Effort Analysis}

\begin{table}[t]
    \center
    \small
    % \scalebox{1.0}{
    \begin{tabular}{l c}
        \toprule
        {\textbf{Model}} & {\textbf{Accuracy(\%)}} \\
        \midrule
        T5-3B parser & $8.19$ \\
        $\our$ & $\mathbf{15.89}$ \\
        \bottomrule
    \end{tabular}
    %}
    \caption{Annotation effort analysis. $\our$ outperforms T5-3B parser trained on samples used for annotating questions $\mathcal{Q}$, justifying our effort for creating questions for $\our$.}
    \label{tbl:ann_effort}
\end{table}

We use 74 samples i.e. one example per slot to design questions for slots.
To analyze the annotation effort, we train an end-to-end parser using these 74 samples and compare it against our method.
We fine-tune T5-3B model~\cite{raffel2020exploring} on these samples with user utterance as input and its corresponding meaning representation as output.
The complete match accuracy on the MTOP dataset is shown in Table~\ref{tbl:ann_effort}.
We observe that $\our$ performs significantly better than the parser trained on these 74 samples, justifying the annotation effort to create questions for $\our$.

\begin{table}[t]
    \center
    \small
    % \scalebox{1.0}{
    \begin{tabular}{l c}
        \toprule
        {\textbf{Strategy}} & {\textbf{Accuracy(\%)}} \\
        \midrule
        Greedy & $15.89$ \\
        Beam search (k=3) & $16.86$ \\
        \bottomrule
    \end{tabular}
    % }
    \caption{Greedy and Beam search prompting strategies are compared. We observe that beam search-based prompting can improve performance when validation data is provided.}
    \label{tbl:greedy_beam}
\end{table}
\subsection{Greedy vs Beam search}
In our proposed method, we employ a greedy strategy where we hierarchically prompt for top-level intent and for its corresponding slots.
We compare it with the beam search strategy with beam size $k$ for every prediction.
% We compare it with the beam search strategy where we consider 3 as the beam size for every prediction.
Specifically, we consider $k$ top-level intents and prompt for their corresponding slots, consider top-$k$ slot values for every slot and finally compute the best meaning representation based on their aggregated NLL scores.
The NLL score of intent $\mathcal{I}_m$, its slots $\mathcal{S}_j \in \Ch(\mathcal{I}_m)$, and their corresponding slot values $\mathrm{slotValues}[\mathcal{S}_j]$ is aggregated as follows:
\begin{equation*}
    \small
    \begin{aligned}
    \alpha \log p(\mathcal{I}_m)+ (1 - \alpha)\sum_{\mathcal{S}_j \in \Ch(\mathcal{I}_m)} \log p(\mathrm{slotValues}[\mathcal{S}_j] | \mathcal{I}_m)
    \end{aligned}
\end{equation*}
where $\alpha$ is tuned on a held-out validation set.
Note that $p(\mathrm{slotValues} | \mathcal{I}_m)$ is computed recursively for its nested intents.
% \subhro{Change this formula to show product over multiple intents / slots.}
The performance comparison is reported in Table~\ref{tbl:greedy_beam}.
We observe that beam search can further improve the performance when validation data is provided.
\section{Conclusion}

In this paper, we explore zero-shot semantic parsing using large language models.
We propose $\our$ that decomposes semantic parsing task into abstractive and extractive QA tasks and leverage LLMs to answer them.
For identifying top-level intent, we view it as an abstractive QA task and propose to generate an answer in an unconstrained fashion and infer the intent label most similar to the generated description.
We demonstrate that current LLMs fail to abstain from prediction for unanswerable questions, which leads to poor performance.
To address this challenge, we train Abstainer using public QA datasets, that is capable of identifying unanswerable questions and abstaining from prediction.
Extensive experiments on the MTOP dataset show significant improvements over pre-trained LLMs.

% \clearpage\newpage
% \section{Ethical Consideration}

\bibliography{anthology,custom}
\bibliographystyle{acl_natbib}

\appendix

\newpage
\section{Appendix}
\label{sec:appendix}

\subsection{QA Datasets for Training Abstainer}
\label{app:qa_datasets}
We use multiple extractive and abstractive QA datasets to generate synthetic unanswerable samples and train Abstainer.
The details about datasets are mentioned in Table~\ref{tbl:qa_datastats}.

\begin{table}[t]
    \center
    \scalebox{0.72}{
    \begin{tabular}{l l r l}
        \toprule
        {\textbf{Type}} & {\textbf{Dataset}} & {\textbf{\# Samples}} \\
        \midrule
        \multirow{6}{*}{Extractive} & Adversarial QA~\cite{bartolo2020beat} & $36000$\\
        & QA-SRL~\cite{he-etal-2015-question} & $8597$ \\
        & DuoRC~\cite{DuoRC} & $186089$ \\
        & ROPES~\cite{Lin2019ReasoningOP} & $14000$ \\
        & SQuADv2~\cite{rajpurkar-etal-2018-know} & $150000$\\
        & Quoref~\cite{Dasigi2019QuorefAR} & $24000$ \\
        \midrule
        \multirow{4}{*}{Abstractive} & ReCoRD~\cite{wang2019superglue} & $121000$ \\
        & DREAM~\cite{sun2019dream} & $10197$ \\
        & QuaRTz~\cite{tafjord2019quartz} & $3864$ \\
        & Tweet-QA~\cite{xiong-etal-2019-tweetqa} & $10692$\\
        \bottomrule
    \end{tabular}
    }
    \caption{Relevant statistics of the QA dataset used to train Abstainer.}
    \label{tbl:qa_datastats}
\end{table}

% This is an appendix.

\end{document}